\documentclass{article}

\usepackage{arxiv}
\usepackage{wrapfig}
\usepackage[utf8]{inputenc} 
\usepackage[T1]{fontenc}    
\usepackage{hyperref}       
\usepackage{url}            
\usepackage{booktabs}       
\usepackage{amsfonts}       
\usepackage{nicefrac}       
\usepackage{microtype}      
\usepackage{lipsum}		
\usepackage{graphicx}
\usepackage{natbib}
\usepackage{doi}
\usepackage{float}
\usepackage{caption}
\usepackage{subcaption}
\usepackage{booktabs,chemformula}
\usepackage{multirow}

\title{TempoNet: Empowering long-term Knee Joint Angle Prediction with Dynamic Temporal Attention in Exoskeleton Control}

\author{$^a$Lyes Saad Saoud and ~$^{b}$Irfan Hussain, \\
	Khalifa University Center for Autonomous and Robotic Systems, Khalifa University, \\
	National Service and Reserve Authority, Khalifa University, \\
 Advanced Research and Innovation Center, Khalifa University, \\Abu Dhabi, United Arab Emirates, P O Box 127788, Abu Dhabi, UAE \\
        \texttt{$^a$lyes.saoud@ku.ac.ae, $^{b}$irfan.hussain@ku.ac.ae}
}

\renewcommand{\undertitle}{Accepted for presentation at the 2023 IEEE-RAS International Conference on Humanoid Robots, Austin, USA, on December 12-14}
\renewcommand{\shorttitle}{TempoNet: Empowering long-term Knee Joint Angle Prediction with Dynamic Temporal Attention in Exoskeleton Control}

\hypersetup{
pdftitle={TempoNet: Empowering long-term Knee Joint Angle Prediction with Dynamic Temporal Attention in Exoskeleton Control},
pdfsubject={q-bio.NC, q-bio.QM},
pdfauthor={Lyes Saad Saoud and Irfan Hussain},
pdfkeywords={Knee Joint Angle Prediction, Deep Learning, Multi-step Forecasting, Biomechanical Analysis, Rehabilitation},
}

\begin{document}
\maketitle

\begin{abstract}
In the realm of exoskeleton control, achieving precise control poses challenges due to the mechanical delay of exoskeletons. To address this, incorporating future gait trajectories as feed-forward input has been proposed. However, existing deep learning models for gait prediction mainly focus on short-term predictions, leaving the long-term performance of these models relatively unexplored.
In this study, we present TempoNet, a novel model specifically designed for precise knee joint angle prediction. By harnessing dynamic temporal attention within the Transformer-based architecture, TempoNet surpasses existing models in forecasting knee joint angles over extended time horizons. Notably, our model achieves a remarkable reduction of 10\% to 185\% in Mean Absolute Error (MAE) for 100 ms ahead forecasting compared to other transformer-based models, demonstrating its effectiveness.
Furthermore, TempoNet exhibits further reliability and superiority over the baseline Transformer model, outperforming it by 14\% in MAE for the 200 ms prediction horizon. These findings highlight the efficacy of TempoNet in accurately predicting knee joint angles and emphasize the importance of incorporating dynamic temporal attention. TempoNet's capability to enhance knee joint angle prediction accuracy opens up possibilities for precise control, improved rehabilitation outcomes, advanced sports performance analysis, and deeper insights into biomechanical research. Code implementation for the TempoNet model can be found in the GitHub repository: https://github.com/LyesSaadSaoud/TempoNet.

\end{abstract}
\begin{keywords}
 ~Exoskeleton control, Gait prediction, Deep learning model, Knee joint angle prediction.
\end{keywords}
\begin{figure}[h]
\centering
\includegraphics[width=0.88\textwidth]{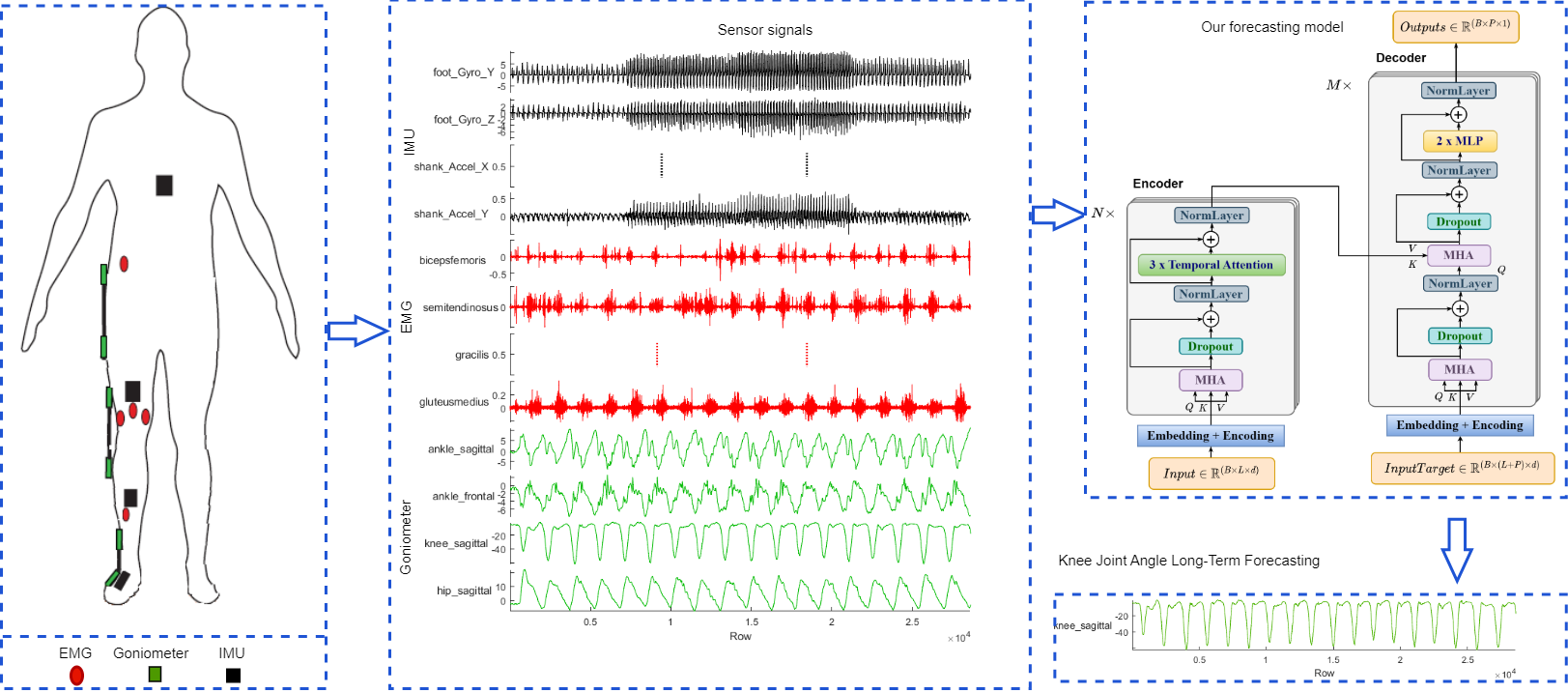}
\caption{Overview of the proposed TempoNet model for long-term gait trajectory prediction in exoskeleton control systems. The model consists of three main components: Dynamic Temporal Attention, Selective Temporal Feature Emphasis, and Enhanced Sequence Modeling. The input-output data obtained from sensors: EMG, acceleration and
gyroscope information from 6-axis IMUs, and joint angle data from electrogoniometers. Dynamic Temporal Attention captures sequential dependencies, Selective Temporal Feature Emphasis emphasizes relevant temporal dependencies, and Enhanced Sequence Modeling models complex relationships using the TempoNet architecture. The output is a predicted long-term gait trajectory for exoskeleton control and rehabilitation applications.}  
\label{main_model}
\end{figure}
\section{INTRODUCTION}

For individuals coping with conditions such as spinal cord injuries, developmental disabilities in children, and neuromotor impairments in adults, the precise detection and classification of lower limb activities have become of utmost importance in accelerating the implementation of active training protocols during the comprehensive rehabilitation process \citep{calabro2016robotic, gonzalez2021robotic, zhou2020accurate}. Exoskeleton robots have emerged as a viable alternative to traditional artificial rehabilitation exercises, showcasing promising initial outcomes \citep{meng2015recent, morone2017robotassisted}. These robots aim to achieve human-like control, which necessitates real-time recognition of human motion intentions and the formulation of effective control strategies to enable seamless interaction and efficient rehabilitation \citep{9863882}. Particularly, wearable exoskeletons have revolutionized gait rehabilitation by offering a unique blend of user involvement and physical activity promotion, enabling their potential use as assistive devices in community settings. Consequently, research on wearable exoskeletons has witnessed a surge in recent years, aligning with the broader trend towards rehabilitation robots \citep{Buchholz2003-ms, Esquenazi2019-wq}. While some exoskeletons have already obtained regulatory approvals and are commercially available, many others are still in the developmental phase \citep{Rodríguez-Fernández2021}.

The increasing attention on wearable robotic exoskeletons transcends various domains, including healthcare, industry, space, and military applications \citep{A1, A2, A3, A4}. In healthcare, their utilization primarily revolves around enhancing the mobility of post-stroke patients and individuals with age-related disorders, facilitating their rehabilitation and providing assistance to physical therapists \citep{A5}. Effective control of exoskeletons is crucial to optimize their performance and interaction with the user \citep{A6}. High-level control, encompassing user intention detection, terrain recognition, and event estimation, plays a pivotal role in enhancing the functionality of exoskeletons \citep{A7, A8}.

Accurate analysis of gait data is essential to achieve effective exoskeleton control as it enables the detection of user intentions and estimation of key gait parameters such as angular positions, velocities, and accelerations. This information serves as the foundation for developing precise controllers that enhance the overall performance of exoskeletons \citep{A9}. Various sensors, including inertial measurement units (IMUs), motion capture systems, foot pressure insoles, gyroscopes, accelerometers, electromyography (EMG), and electroencephalography (EEG), are employed to estimate kinetic and kinematic parameters, as well as muscle and brain activities. Researchers have devised multiple algorithms to analyze and process sensor data, evaluate human intention, and predict essential gait information \citep{A9}.

Machine learning and deep learning algorithms are extensively employed in a wide range of classification and regression tasks due to their capacity to extract high-level abstract features from extensive datasets \citep{9863882}. These algorithms excel in establishing accurate relationships between inputs and outputs in nonlinear systems, exhibiting enhanced capability in handling data variability, particularly in lower-limb locomotion and pathological gait analysis \citep{A10, A11}. Deep learning models, including Transformers \citep{vaswani}, have emerged as prominent tools in gait analysis, surpassing traditional machine learning methods such as Support Vector Machines (SVM) and Multilayer Perceptron (MLP) \citep{A17}.

The inherent mechanical delay of exoskeletons presents a significant challenge in achieving precise control, as it can impact the system's response time \citep{exoskeleton_mechanical_delay}. Given the requirement for accurate and timely responses, a potential solution lies in incorporating feed-forward input comprising future gait trajectories. Existing literature on deep learning models for gait prediction has primarily focused on short-term predictions with limited output window sizes \citep{LSTM_Angle, LSTM_CNN_Angle, LSTM_acceleration1, LSTM_acceleration2}. Consequently, the long-term performance of deep learning models in gait prediction remains an area that warrants exploration.

To address these limitations, this study proposes an enhanced variant of the Transformer model specifically tailored for predicting knee joint angles from time series data (see Fig. \ref{main_model}). The proposed model incorporates dynamic temporal attention to capture sequential dependencies, utilizes selective temporal feature emphasis to highlight relevant temporal patterns, extends the receptive field of attention mechanisms, and develops techniques to handle variable-length knee joint angle time series effectively. By implementing these modifications, the proposed model aims to provide a more comprehensive understanding of the sequential information inherent in knee joint angle time series.

The research gap identified in this study is the lack of an effective and accurate deep learning model specifically designed for long-term gait trajectory prediction in exoskeleton control systems. Existing deep learning models, including the traditional Transformer, have not been adequately explored for this purpose. Furthermore, the shortcomings of the traditional Transformer architecture in handling time series data, such as the limited receptive field, fixed positional encodings, and inability to accommodate variable-length inputs, highlight the need for enhanced variants to achieve more accurate knee joint angle predictions.

The key contributions and originality of our model are as follows:

\begin{itemize}
\item Dynamic Temporal Attention: Unlike traditional models that employ static attention mechanisms, our model introduces dynamic temporal attention. This innovative approach allows the model to focus on relevant temporal patterns dynamically and capture time-dependent relationships within the input data. By adaptively attending to different time steps, our model enhances its ability to capture crucial temporal dependencies and improve prediction accuracy.

\item Selective Temporal Feature Emphasis: Our model utilizes learnable linear projections to emphasize specific time steps selectively, highlighting relevant temporal dependencies and features. This selective attention mechanism improves the model's ability to capture critical temporal patterns, leading to more accurate knee joint angle predictions.

\item Enhanced Sequence Modeling: Our model builds upon the powerful sequence modeling capabilities of the Transformer architecture. By leveraging the attention mechanism and position-wise feed-forward network in the encoder layer, our model captures complex relationships within the input sequence. This facilitates accurate prediction of knee joint angles by effectively modeling the underlying dynamics and dependencies.

\item Improved Performance: Our model outperforms traditional models in multistep knee joint angle prediction tasks, demonstrating higher accuracy and prediction quality. It excels in capturing and modeling the intricate temporal patterns of knee joint angles, leading to improved predictions.
\end{itemize}
\begin{figure*}[h]
\centering
\includegraphics[width=0.9\textwidth]{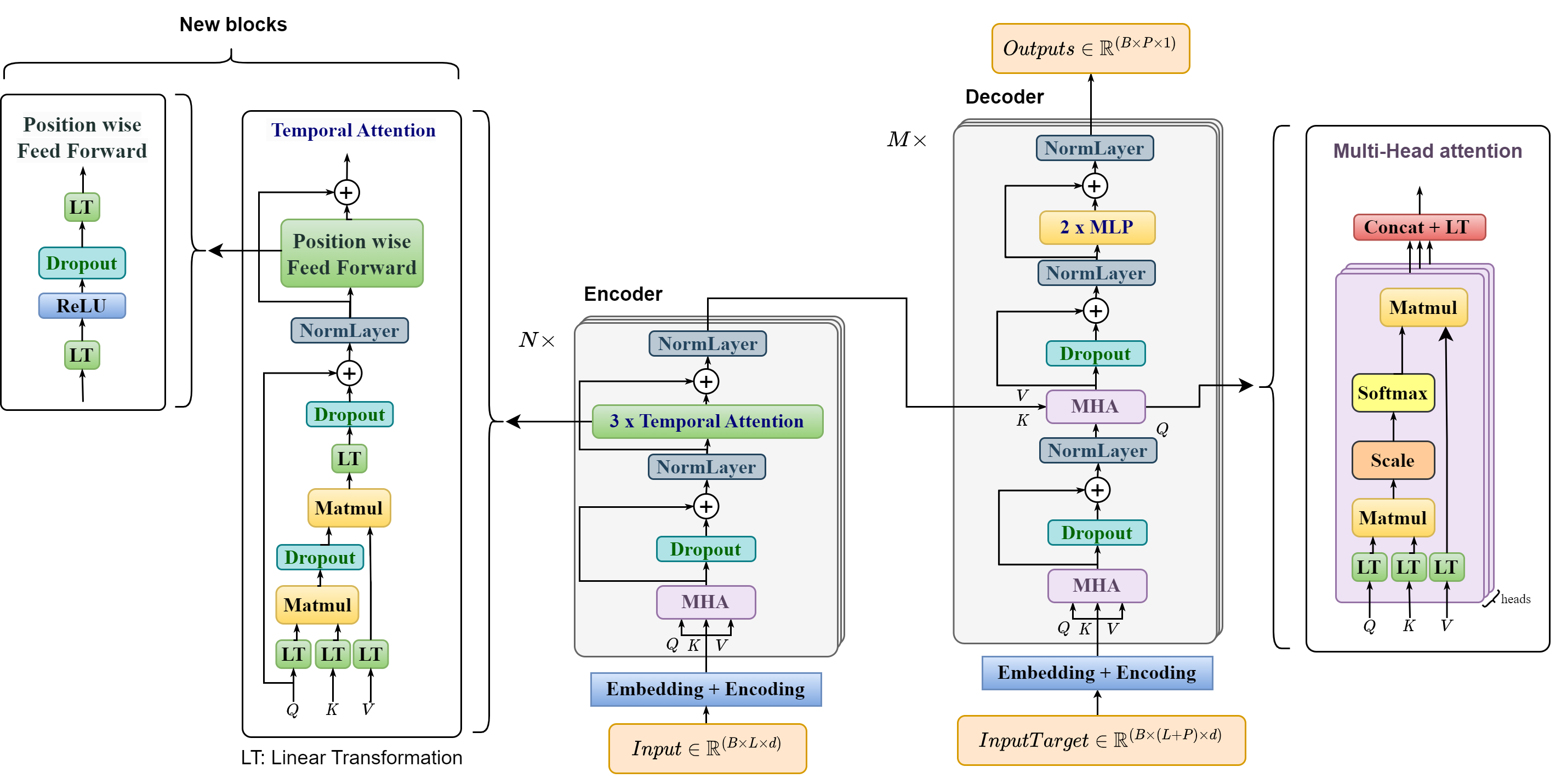}
\caption{TempoNet architecture. The TempoNet consists of N encoders and M decoders. The decoder block is the same as the original Transformer model. The encoder has three temporal attention blocks in place of multilayer perceptrons with an incorporated position-wise feed-forward block. $Input$ is the sensor data and $Output$ is the forecasted sagittal knee angle.}  
\label{model_architecture}
\end{figure*}

The rest of this paper is organized as follows: Section II provides a concise review of related works. Section III presents the proposed TempoNet model architecture, detailing the Dynamic Temporal Attention. Section IV describes the experimental setup and dataset used for evaluating the proposed model's performance. Section V presents the results and discussion, analyzing the experimental findings and exploring potential applications of the proposed model in the field of wearable robotics and exoskeletons. Finally, Section VI concludes the paper by summarizing the key findings and outlining future research directions.

\section{Related Work}

In recent years, numerous studies have focused on gait analysis and prediction using deep learning models. Gait Prediction Transformer  explored the stability of deep learning models in one-step-ahead gait trajectory prediction \citep{Gait_Prediction_Transformer}. Their findings revealed that the Transformer model exhibited greater resilience to noise compared to traditional deep learning models like LSTM \citep{LSTM} and CNN \citep{CNN}, while maintaining comparable accuracy. However, the study primarily focused on short-term predictions, and the long-term performance of deep learning models in gait prediction for exoskeleton control systems remains unexplored.

Several approaches have been proposed to address the limitations of the traditional Transformer architecture in time series forecasting. LogTrans \citep{li2020enhancing} introduced local convolution to enhance the Transformer's capability to capture temporal dependencies. Reformer \citep{kitaev2020reformer} employed local-sensitive hashing (LSH) attention to reduce the computational complexity of the Transformer, making it more suitable for long-term time series forecasting. Informer \citep{haoyietal-informer-2021, haoyietal-informerEx-2023} introduced KL-divergence-based ProbSparse attention to capture both short-term and long-term temporal patterns effectively.

In addition to modified Transformer architectures, other deep learning models have been applied to time series forecasting. DeepAR \citep{SALINAS20201181} combines autoregressive methods, recurrent neural networks (RNNs), and convolutional neural networks (CNNs) to model future series with probabilistic distributions. LSTNet \citep{lai2018modeling} combines CNNs and RNNs to capture short-term and long-term temporal patterns, respectively.

Attention-based RNNs have been introduced to explore long-range dependencies in time series data. Temporal convolution networks (TCN) \citep{bai2018empirical} employ causal convolutions to model temporal causality effectively. These models have shown promising results in various time series forecasting tasks.

Furthermore, time series-based decomposition techniques have been widely used in time series analysis. Prophet \citep{prophet}, N-BEATS \citep{oreshkin2020nbeats}, and DeepGLO \citep{sen2019think} employ trend-seasonality decomposition, basis expansion, and matrix decomposition, respectively. These decomposition techniques are valuable for exploring historical changes over time. However, they may overlook the hierarchical interaction between underlying patterns in the long-term future. To address this issue, Wu et al. proposed the Autoformer model \citep{wu2021autoformer}, which enables progressive decomposition throughout the forecasting process, considering both past series and predicted intermediate results. Recently, FocalGatedNet model has been proposed, which incorporated the dynamic contextual focus Attention and gated linear units to enhance feature dependencies and interactions, providing a new and promising approach for accurate knee joint angle prediction \citep{ibrahim2023focalgatednet}.

While these existing methods have contributed to the field of gait analysis and time series forecasting, there is a research gap in developing an effective and accurate deep learning model specifically designed for long-term gait trajectory prediction in exoskeleton control systems. The proposed TempoNet model addresses this gap by introducing dynamic temporal attention, selective temporal feature emphasis, and enhanced sequence modeling. It overcomes the limitations of the traditional Transformer architecture and other existing models, leading to improved knee joint angle predictions for exoskeleton control and rehabilitation applications.

The inherent mechanical delay of exoskeletons \citep{exoskeleton_mechanical_delay} presents a significant challenge in achieving precise control, as it can impact the system's response time. Given the requirement for accurate and timely responses, a potential solution lies in incorporating feed-forward input comprising future gait trajectories. Existing literature on deep learning models for gait prediction has primarily focused on short-term predictions with limited output window sizes  \citep{LSTM_Angle, LSTM_CNN_Angle, LSTM_acceleration1, LSTM_acceleration2}. Consequently, the long-term performance of deep learning models in gait prediction remains an area that warrants exploration.

To address these identified weaknesses, this research proposes an enhanced variant of the Transformer model specifically tailored for predicting knee joint angles from time series data. By incorporating temporal attention to capture sequential dependencies, improving positional encodings to convey dynamic positional information, extending the receptive field of attention mechanisms to capture long-range dependencies, and developing techniques to handle variable-length knee joint angle time series effectively, the proposed model aims to provide a more comprehensive understanding of the sequential information inherent in knee joint angle time series.

\section{The proposed TempoNet model}
TempoNet is a gait prediction model that extends the Transformer architecture. It consists of multiple encoder blocks and decoder blocks, as shown in Fig. \ref{model_architecture}. The encoder block incorporates a modified attention mechanism designed specifically for gait prediction tasks, while the decoder block follows the standard Transformer decoder architecture.

In the TempoNet encoder, the input sensor data is embedded into a continuous representation and then processed through a series of self-attention and temporal attention blocks. The self-attention mechanism allows the model to capture dependencies between different time steps in the input sequence, while the temporal attention layers help in learning complex temporal patterns.
The dynamic temporal attention block in TempoNet allows the model to dynamically attend to different temporal patterns within the input sequence. The encoder's multi-head attention (MHA) block uses complete self-attention, allowing each attention operation to pay attention to the whole input sequence. It operates by computing attention scores between the query ($Q$) and key ($K$) vectors, and then using the attention scores to obtain weighted sums of the value ($V$) vectors.
Given an input sequence $x\in\mathbb{R}^{B\times L\times d}$, where $B$ is the batch size, $L$ is the sequence length, and $d$ is the size of the hidden input embedding, the self-attention mechanism computes the query, key, and value representations as follows:
\begin{equation} \label{eq1}
\begin{split}
Q = x \; W_Q &\in \mathbb{R}^{B \text{x} L \text{x} d} \\ 
K = x \; W_K &\in \mathbb{R}^{B \text{x} L \text{x} d} \\
V = x \; W_V &\in \mathbb{R}^{B \text{x} L \text{x} d} 
\end{split}
\end{equation}
where $\text{W}_\text{Q}, \text{W}_\text{K}, \text{W}_\text{V}$ are linear projections, each of size $d\times d$.

Then, the dot-product attention scores between queries and keys are computed as:
\begin{equation} \label{eq2}
    A = Softmax \left(\frac{QK^T}{\sqrt{dh}}\right)
\end{equation}
where $h$ is the number of attention heads, $A$ is the attention weights tensor, and $\frac{1}{\sqrt{dh}}$ is the scaling factor SF.

If a mask is provided, it is applied to the attention scores as:
\begin{equation} \label{eq3}
    A = A \odot M
\end{equation}
where $M\in {0,1}^{B\times h\times L\times L}$ is the mask tensor.

Next, another dot-product multiplication between the resulting output $A$ matrix and values is performed as:
\begin{equation} \label{eq4}
    C =  AV^T
\end{equation} 
The output $C$ was then normalized and passed through our new temporal attention blocks. 
Let $\mathbf{X}$ represent the normalized sequence output $C$. The dynamic temporal attention block consists of the following steps:

The input sequence $\mathbf{X}$ is projected into query ($Qt$), key ($Kt$), and value ($Vt$) vectors using linear transformation operations:
\begin{equation} \label{eq5}
    Qt = \mathbf{X} \cdot \mathbf{Wt}_Q, \quad Kt = \mathbf{X} \cdot \mathbf{Wt}_K, \quad Vt = \mathbf{X} \cdot \mathbf{Wt}_V
\end{equation} 

where $\mathbf{Wt}_Q \in \mathbb{R}^{d \times d}$, $\mathbf{Wt}_K \in \mathbb{R}^{d \times d}$, and $\mathbf{Wt}_V \in \mathbb{R}^{d \times d}$ are learnable weight matrices.

The attention scores are computed by taking the dot product between the query and key vectors:
\begin{equation} \label{eq6}
    \text{scores} = Qt \cdot Kt^T
\end{equation}

The attention scores are normalized using the softmax function to obtain attention weights:

\begin{equation} \label{eq7}
   \text{weights} = Softmax(\text{scores})
\end{equation}

The attention weights are applied to the value vectors to obtain a weighted sum, representing the attended information in the input sequence:

\begin{equation} \label{eq7}
   \text{attended\_output} = \text{weights} \cdot Vt
\end{equation}

The attended output contains the relevant information from the input sequence, which captures the dynamic temporal dependencies.

These steps allow the model to dynamically attend to different temporal patterns within the input sequence and focus on the most relevant information for the task at hand.
This process is repeated for multiple encoder blocks, allowing the model to learn hierarchical representations of the input sequence.

The decoder's design is largely identical to that of the encoder, with the exception of the cross-attention MHA block, which comes before the MLP and the use of MLPs instead of temporal attention. Similar to the other MHA blocks, the cross-attention module also accepts encoder outputs as inputs for the keys and values. Additionally, masking is used in the decoder's initial MHA block to stop information from flowing from later places.

\section{Methodology}
\subsection{Dataset}

This work uses an open-source dataset on human gait kinematics and kinetics in four different locomotion activities\citep{dataset}. 22 healthy individuals wearing various sensors carried out the following locomotion tasks in the dataset: walking on level ground at a slow, normal, and fast pace compared to the subject's preferred speed on clockwise and counterclockwise circuits; on a treadmill at 28 speeds ranging from 0.5 m/s to 1.85 m/s in 0.05 m/s increments; and up and down a ramp with inclines of 5.2$^{\circ}$, 7.8$^{\circ}$, 9.2$^{\circ}$, 11$^{\circ}$, 12.4$^{\circ}$, and 18$^{\circ}$; and up and down stairs with step heights of 4 in, 5 in, 6 in, and 7 in.

The data collection contains three wearable sensor signals: EMG data from distinct muscle groups, acceleration and gyroscope information from 6-axis IMUs, and joint angle data from electrogoniometers.
The EMG data were taken at a sample frequency of 1000 Hz and bandpass filtered at a cutoff frequency of 20 Hz - 400 Hz. It consists of 11 muscle groups: gluteus medius, external oblique, semitendinosus, gracilis, biceps femoris, rectus femoris, vastus lateralis, vastus medialis, soleus, tibialis anterior, and gastrocnemius medialis. Four IMUs on the torso, thigh, shank, and foot segments were collected at a frequency of 200 Hz and lowpass filtered at 100 Hz. Three GONs situated on the hip, knee, and ankle were recorded at 1000 Hz and filtered with a lowpass filter with a cutoff frequency of 20 Hz.

The data set offers processed biomechanics data in addition to the sensor's raw data, such as inverse kinematics/dynamics, joint power, gait cycle, force plate, and motion capture data.



\subsection{ Experimental Setup and Model Configuration}

We used a workstation equipped with a GPU NVIDIA GeForce RTX 4090 and  PyTorch 1.13.1 to train the models. The Adam optimizer with a learning rate of $10^{-4}$ was utilized. The training was performed for 10 epochs, and an early stopping algorithm was employed to prevent overfitting. The batch size was set to 32. To account for the stochastic nature of training, the models were trained for 10 iterations, and the best results were selected. A single subject's data were used, divided into 80\% for training and 20\% for testing. 
Inference time was measured by running the models 1000 times, with 100 warm-up steps for the GPU, and recording the average time for a single step.
The input features consisted of EMG, GON, and IMU sensor data, resulting in a total of 40 features. The output variable of interest was the knee sagittal angle. All data were normalized to achieve zero mean and unit standard deviation.

To align the sampling rates, the IMU and toe/heel events were up-sampled using interpolation from 5 ms to 1 ms intervals, matching the rates of the EMG and goniometer data. The lookback window was set to 128 ms (128 data points), and various forecasting horizons (ranging from 1 ms to 100 ms) were considered. These horizons aimed to investigate the delay between EMG sensor muscle activation and the subsequent force and movement production, as discussed in \citep{viitasalo}. Additionally, longer prediction times aimed to compensate for the response delay caused by the mechanical components of the exoskeleton \citep{exoskeleton_mechanical_delay}.

The performance of TempoNet will be compared to other recent state-of-the-art Transformer-based models, including LSTM \citep{HochSchm97}, Transformer \citep{vaswani}, Autoformer \citep{autoformer}, Informer \citep{informer}, DLinear, and NLinear \citep{LTSF}. The Mean Absolute Error (MAE) metric is the metric used to evaluate the models' performance.

TempoNet comprises 4 encoder layers and 3 decoder layers, as these settings yielded optimal results for our specific case. Interestingly, superior results were achieved by omitting temporal embeddings in TempoNet (using only value + positional embedding), while the other models performed best without positional embeddings (using only value + temporal embedding). The remaining transformer-based models will retain their default number of layers (2 encoders, 1 decoder), and the stacked LSTM model will consist of 1 encoder/decoder layer with a kernel size of 3. The number of attention heads (8 heads) and model dimensions (512 for the model and 2048 for the feed-forward network) will remain consistent. Despite having more encoder and decoder layers, TempoNet exhibited improved inference time compared to most other transformer-based models, as demonstrated in the subsequent section.

\section{Results and Discussion}

This section presents a comprehensive comparative analysis of different models for predicting sagittal knee angles across various prediction lengths. Our aim is to explore and evaluate the performance of these models in a rigorous manner to provide insights into their effectiveness. The results of our analysis are summarized in Table \ref{Results}.

The findings of our study reveal intriguing patterns and distinctions among the models. Specifically, we observe that the LSTM model exhibits remarkable performance when it comes to shorter prediction lengths. Its ability to effectively capture short-term dependencies and patterns allows it to outperform other models in this context. On the other hand, the TempoNet model demonstrates superior performance in longer-term predictions, particularly within the 20-100 ms range. It surpasses the performance of other models and even outperforms the base transformer model, showcasing its prowess in capturing complex patterns and dependencies for extended forecasting horizons.

\begin{figure}[t]
    \centering
    \includegraphics[scale=0.6]{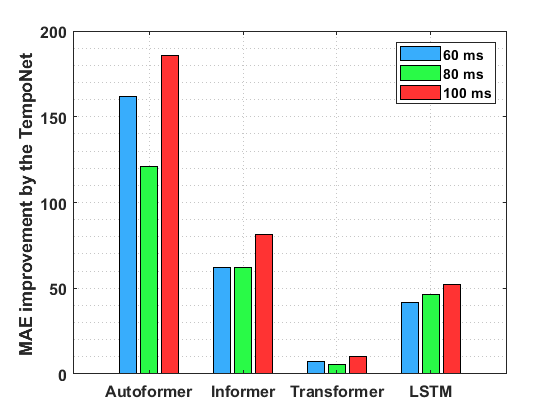}
    \caption{The relative improvement in MAE provided by TempoNet compared to the four studied deep learning models for different forecasting horizons.}
    \label{fig:TempoNetComapere}
\end{figure}

\begin{figure}[ht]
    \centering
    \includegraphics[scale=0.6]{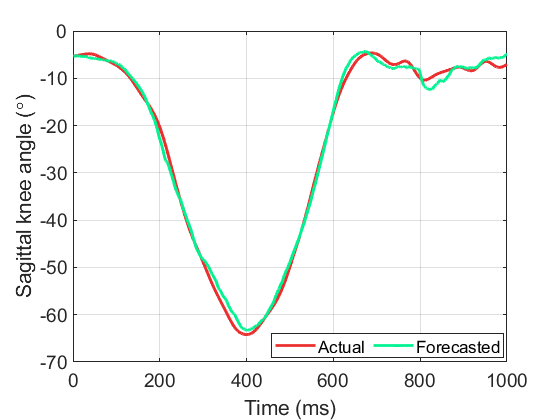}
    \caption{A sample of the forecasting results for a 60-step horizon using the proposed TempoNet is shown.}
    \label{fig:predict_60}
\end{figure}

\begin{figure}[t]
    \centering
    \includegraphics[scale=0.6]{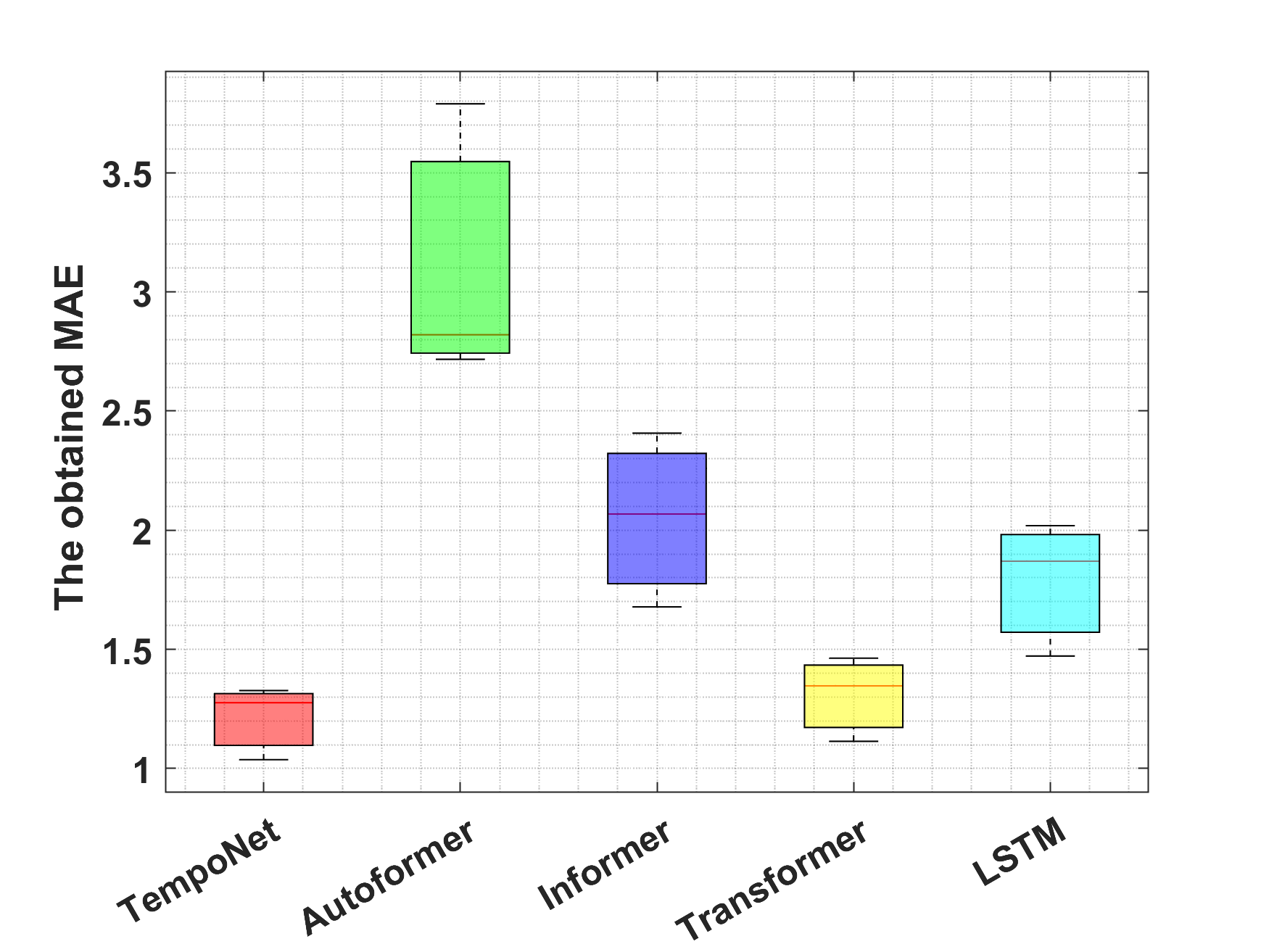}
    \caption{Comparison of MAE across Deep Learning-Based Models for Knee Joint Angle Prediction.}
    \label{fig:box_plot}
\end{figure}

To delve deeper into the performance of the TempoNet model, we examine its specific improvements over the baseline and other models (see Fig. \ref{fig:TempoNetComapere}). To provide a comprehensive representation of the results and gain a holistic understanding, we present a box plot (Fig. \ref{fig:box_plot}) illustrating the MAE obtained from all deep learning-based models across various horizon time scales. Analyzing the box plot, it becomes evident that TempoNet exhibits a compact box, as well as the lowest average and maximum values among the compared models. These compelling results position TempoNet as a highly promising model for accurate knee joint angle prediction. For instance, at a prediction length of 60 ms, TempoNet achieves a significant 7.43\% reduction in MAE compared to the baseline (see Fig. \ref{fig:predict_60}). Similarly, at a prediction length of 80 ms, TempoNet exhibits an impressive decrease in MAE by 121\%, 62\%, 5\%, 46\%, 737\%, and 587\% compared to the Autoformer, Informer, Transformer, LSTM, DLinear, and Non-Linear, respectively. Furthermore, at a prediction length of 100 ms, TempoNet shows substantial improvements, with 185\%  81\%  10\%   52\%  877\%  722\% decrease in MAE compared to all models presented in Table \ref{Results}. These results highlight the ability of the TempoNet model to refine and enhance predictions as the prediction length increases.

\begin{table*}[t]
\footnotesize
\caption{The obtained MAE for forecasting sagittal knee angle using different models. The best results are highlighted in bold..}
\label{Results}
\centering
\begin{tabular}{|c|c|c|c|c|c|c|c|} 
 
\hline

\multicolumn{1}{|c|}{Pred. Length}    
& \multicolumn{1}{c|}{TempoNet}                       
& \multicolumn{1}{c|}{Autoformer}                      
& \multicolumn{1}{c|}{Informer}  
& \multicolumn{1}{c|}{Transformer}   
& \multicolumn{1}{c|}{LSTM}
& \multicolumn{1}{c|}{DLinear}                   
& \multicolumn{1}{c|}{Non-Linear} \\ 
\hline
1& 0.442  & 1.548   & 0.608    & 0.288  & \textbf{0.045}    & 0.225    & 0.147   \\ 
20& \textbf{0.670} & 1.800   & 0.821 & 0.705  & 0.967 & 3.144   & 2.438    \\
40& \textbf{0.815}    & 2.121  & 0.990  & 0.819 & 1.339   & 5.997  & 4.722    \\
60& \textbf{1.037}     & 2.717   & 1.678  & 1.114  & 1.472  & 8.360  & 6.749   \\
80& \textbf{1.276} & 2.820  & 2.068 & 1.347 & 1.870   & 10.688 & 8.772    \\
100 & \textbf{1.327}    & 3.789  & 2.407 & 1.463   & 2.019  & 12.973  & 10.917  \\
\hline
\end{tabular}
\end{table*}

\begin{table}[t]
\caption{Model Parameters and Average Running Time}
\label{eff_table}
\centering
\begin{tabular}{|c|c|c|c|}
\hline
Model         & Params. & Train Time 
& Infer. Time  \\ \hline   
TempoNet 
& $\sim$71.59 M*    & $\sim$1400 s                
& 7.47 $\pm$ 0.73 s                              \\ 
Autoformer    
& $\sim$10.71 M    & $\sim$1800 s                  
& 19.04 $\pm$ 1.76 s                              \\

Informer      
& $\sim$11.45 M    & $\sim$1300 s
& 6.00 $\pm$ 1.02 s  \\

Transformer   
& $\sim$10.66 M    & $\sim$900 s             
& 2.76 $\pm$ 0.66 s                              \\

LSTM   
& $\sim$8.11 M    & $\sim$1200 s             
& 7.37 $\pm$ 0.50 s                             \\ 

DLinear       
& $\sim$25.8 K     & $\sim$80 s             
& 0.03 $\pm$ 0.00 s                              \\
NLinear       
& $\sim$12.9 K     & $\sim$80 s                    
& 0.01 $\pm$ 0.00 s                              \\ \hline

\multicolumn{4}{l}{\footnotesize \emph{*Increase is due to using more encoder/decoder layers}}
\end{tabular}
\end{table}

While TempoNet excels in longer prediction lengths, it is important to acknowledge instances where other models may outperform it. For instance, at a prediction length of 1 ms, the Dlinear, Nlinear, LSTM, and baseline Transformer models showcase superior performance. However, as we extend the prediction lengths to 20 ms, 40 ms, 60 ms, 80 ms, and 100 ms, TempoNet consistently outperforms the other models. On average, it demonstrates a remarkable 62\% improvement in MAE compared to all models, making it the top-performing model in our analysis. These results underscore the superior long-term predictive capabilities of the TempoNet model in accurately estimating knee angles.

Additionally, to assess the robustness of our model, we conducted further experiments comparing the Mean Absolute Error (MAE) of TempoNet with the baseline Transformer model, which provided the closest results, for a forecasting horizon of 200 ms. The MAE results for the 200 ms prediction length are 2.515$^{\circ}$ for TempoNet and 2.861$^{\circ}$ for the baseline Transformer model.
These findings indicate that TempoNet outperforms the baseline Transformer model by 14\% in terms of MAE for the 200 ms prediction horizon, further demonstrating its reliability and superior performance in long-term knee joint angle prediction tasks. These impressive results highlight TempoNet as a robust and promising choice for accurate knee angle estimation over extended forecasting horizons.
In terms of computational efficiency, our model demonstrates a notable advantage in both training and inference time when compared to the Autoformer model, which represents the latest advancement among the competing models, as depicted in Table \ref{eff_table}. The overall inference time remains relatively consistent across the selected lengths for both Transformer-based models and linear models. However, it is important to note that LSTM's inference time exhibits a substantial increase as the output prediction length extends, ranging from 8 seconds for a single time-step to 20 seconds for 100-time steps. This can be attributed to LSTM's sequential information processing approach, which becomes increasingly challenging for longer sequences and leads to significant computational overhead.

In discussing computational complexity, it is important to highlight that our analysis primarily concentrates on relatively small output lengths, where the quadratic time and memory complexity ($\mathcal{O}(L^2)$) associated with transformers do not significantly affect our findings. However, for larger output lengths, it has been observed that the base transformer model outperforms the Informer and Autoformer models at lower prediction lengths \citep{autoformer}. Nevertheless, as the prediction length increases exponentially, the transformer model's memory requirements become more demanding, while the Autoformer and Informer models maintain a more reasonable memory footprint and efficiency of $\mathcal{O}(L \log L)$.

Based on our analysis, we identify the optimal prediction interval for knee angle prediction to fall between 20 ms and 100 ms. This observation aligns with previous research and recommendations \citep{viitasalo}, which claims that the EMG signals are produced with a delay of 30-100 ms between the activation of muscles, as detected by the EMG sensor, and the force and movement generated by the muscle.

Despite TempoNet's higher performance in predicting knee joint angles across lengthy time horizons, it is essential to recognize the need for more testing on other datasets. Testing the model on additional datasets with various properties, such as demographics, activity kinds, and exoskeleton designs, should be part of future studies. This will increase TempoNet's generalizability and demonstrate its efficacy in many contexts.

The incorporation of dynamic temporal attention in the current TempoNet implementation may need the use of substantial computing resources. Future studies might investigate optimization methods to simplify the model's computations and increase its suitability for real-time applications in order to solve this issue. This might entail using model pruning, model compression, or model quantization techniques to speed up inference without sacrificing accuracy.

Although TempoNet's findings are encouraging, it is essential to assess their effectiveness in actual exoskeleton users' real-world deployment settings. It would be helpful to run tests and get information from people using wearable exoskeletons in order to better understand how useful and practical the concept is. Further enhancing the model's robustness and dependability would include gathering user input and addressing difficulties discovered during real-world deployment.

Future research can improve TempoNet's applicability, effectiveness, and generalizability by resolving these drawbacks, making it more dependable for use in exoskeleton control and rehabilitation applications.

\section{CONCLUSIONS}
This study introduces TempoNet, a novel deep learning model that incorporates dynamic temporal attention and selective temporal feature emphasis into the standard Transformer architecture, showcasing remarkable performance in predicting sagittal knee angles. Through extensive experimentation on a comprehensive dataset, TempoNet surpasses various cutting-edge models, including Transformer, Autoformer, Informer, NLinear, DLinear, and LSTM.

A noteworthy achievement of TempoNet is its significant advancements in long-term prediction, substantially enhancing the accuracy of knee joint angle estimation. Notably, our model achieves a reduction of 10\% to 185\% in Mean Absolute Error (MAE) for 100 ms ahead forecasting compared to other transformer-based competitors. Furthermore, TempoNet demonstrates further reliability and superiority over the baseline Transformer model, outperforming it by 14\% in MAE for the 200 ms prediction horizon.

These compelling results underscore the significance of the proposed modifications tailored specifically for predicting knee joint angles, particularly in long-term scenarios. The findings from TempoNet's research offer promising prospects for future investigations into wearable exoskeletons and rehabilitation techniques targeting individuals with neurological impairments.



\section*{Acknowledgment}
This work was supported in part by the Advanced Research and Innovation Center (ARIC), which is jointly funded by Mubadala UAE Clusters and Khalifa University of Science and Technology, and in part by Khalifa University Center for Autonomous and Robotic Systems under Award RC1-2018-KUCARS.

\bibliographystyle{chicago}
\bibliography{bibliography}

\end{document}